\pdfoutput=1
\documentclass[10pt,twocolumn,letterpaper]{article}

\usepackage{cvpr}
\usepackage{times}
\usepackage{epsfig}
\usepackage{graphicx}
\usepackage{amsmath}
\usepackage{amssymb}
\usepackage{multirow}
\usepackage{enumitem}
\usepackage{xcolor,colortbl}

\usepackage[breaklinks=true,letterpaper=true,colorlinks,bookmarks=false]{hyperref}
\newcommand{\mytilde}{\raise.17ex\hbox{$\scriptstyle\mathtt{\sim}$}}

\cvprfinalcopy

\title{Low-level Vision by Consensus in a Spatial Hierarchy of Regions}
\author{Ayan Chakrabarti$^{1}$, Ying Xiong$^{2}$, Steven J. Gortler$^{2}$, Todd Zickler$^{2}$\\
\parbox{18em}{\centering$^{1}$TTI-Chicago}~~~
\parbox{16em}{\centering$^{2}$Harvard University}\\
{\tt\small ayanc@ttic.edu, yxiong@seas.harvard.edu, sjg@cs.harvard.edu, zickler@seas.harvard.edu}
}

\begin{document}
\maketitle

\begin{abstract}
We introduce a multi-scale framework for low-level vision, where the goal is estimating physical scene values from image data---such as depth from stereo image pairs. The framework uses a dense, overlapping set of image regions at multiple scales and a ``local model,'' such as a slanted-plane model for stereo disparity, that is expected to be valid piecewise across the visual field. Estimation is cast as optimization over a dichotomous mixture of variables, simultaneously determining which regions are inliers with respect to the local model (binary variables) and the correct co-ordinates in the local model space for each inlying region (continuous variables). When the regions are organized into a multi-scale hierarchy, optimization can occur in an efficient and parallel architecture, where distributed computational units iteratively perform calculations and share information through  sparse connections between parents and children. The framework performs well on a standard benchmark for binocular stereo, and it produces a distributional scene representation that is appropriate for combining with higher-level reasoning and other low-level cues.

\end{abstract}

\section{Introduction}

\begin{figure*}[!t]
  \centering
  \includegraphics[width=\textwidth]{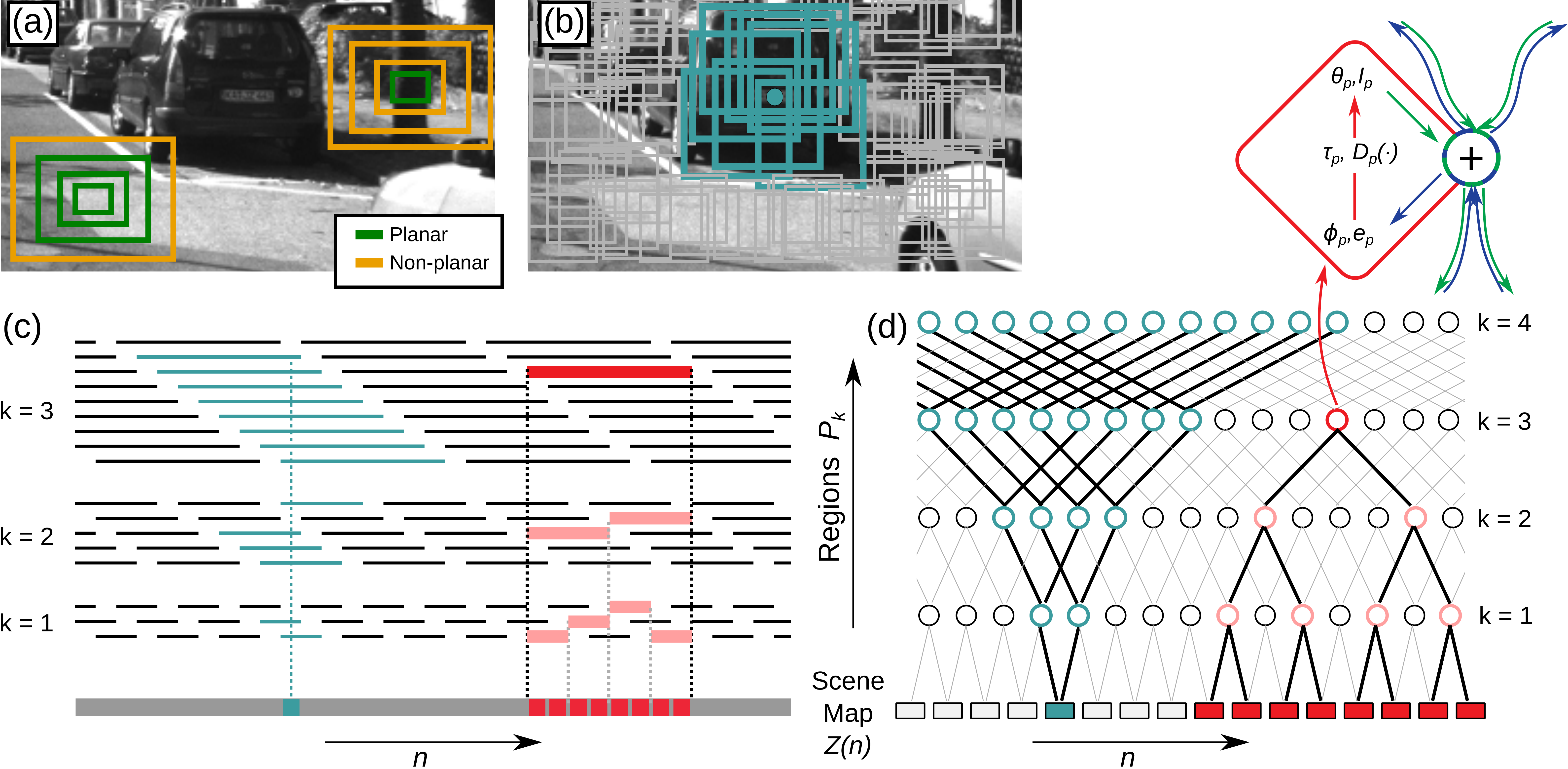}
  \caption{Consensus framework for low-level vision, using binocular stereo as an example. (a) Window-based stereo matching with a slanted-plane model reduces ambiguity, but it requires guessing the correct window shapes and sizes throughout the scene. Consensus addresses this by explicitly considering all regions at all locations (depicted as a 2D cartoon (b) and in 1D organized by scale (c)). It  reasons simultaneously about which regions are inliers to the slanted-plane model and the correct slanted plane for each inlying region. The regional slanted planes must agree where they overlap, and in the objective this implies high-order factors that link the variables of thousands of regions that overlap each pixel (blue in (b) and (c)). When regions are organized hierarchically (red/pink in (b)), optimization becomes parallel and efficient. (d) The result is a distributed architecture, with computational units that iteratively perform the same set of computations and share information sparsely between parents and children. The framework can be applied to a variety of low-level tasks using a variety of regional models.} 
  \label{fig:fwork}
\end{figure*}

Low-level vision is the estimation of depth, motion, shape, and other physical scene properties from visual measurements. Since it is ill-posed, methods often employ a \emph{local model} that is expected to apply piecewise across the scene, and that restricts the variation of scene values within each applicable piece or region. Slanted planes for binocular disparity, affine optical flows, and families of smooth shapes for surface normals are common examples. The restriction on scene variability in applicable regions allows image cues to be aggregated spatially across each region, thereby reducing the ambiguity that exists point-wise.  The fundamental challenge lies in identifying---automatically from the image input---the sizes and shapes of the aggregation regions that are right for each part of a scene. Regions that are too small do not sufficiently reduce the underlying ambiguity, while those that are too big or the wrong shape span abrupt scene changes that violate the local model and make estimates unreliable (\eg, Fig.~\ref{fig:fwork}~(a)).

We introduce a computational framework to address this challenge. Called the \emph{consensus framework}, we apply it to the binocular stereo problem while also presenting it generally as a way to attack a variety of low-level tasks. The framework explicitly considers a large set of dense, overlapping regions of many sizes that redundantly cover the image plane (Fig.~\ref{fig:fwork}~(b)). It simultaneously determines which regions are inliers to the local model (binary variables) and, for each inlying region, the correct coordinates in the local model space for that region (continuous variables). Estimation is cast as optimizing an objective that requires each inlying region to be supported by its local image data while also having scene estimates that are consistent with its overlapping neighbor regions. The output of the framework---the inlier statuses of all regions and the local estimates from the inliers---offers a rich, multi-scale representation of the physical scene. This includes spatial grouping information, a global scene map, and a point-wise measure of confidence, all of which are desirable when seeking to combine multiple low-level cues or integrate higher-level processes.

Compared to traditional approaches based on Markov random fields (MRFs), the consensus framework reasons in a much larger variable space, and more critically, with orders of magnitude more links between variables.  This is because it enforces simultaneous consistency between the thousands of regions that overlap any single pixel. Despite this complexity, two properties make estimation not only feasible, but efficient. First, since the dense region-set embodies an over-complete scene representation---with many more internal variables than values in the output scene map---good solutions can often be reached by a simple alternating algorithm similar to expectation-maximization. Second, we show analytically that when the regions are organized hierarchically by scale (\eg Fig.~\ref{fig:fwork}~(c)), each region only needs to sum information from its parents and children (Fig.~\ref{fig:fwork}~(d)). This leads to a significant reduction in computation because the hierarchical connections constitute only a minuscule fraction of the total links that exist in the consensus objective.

The estimation architecture ends up being composed of a large network of computational units, one for each region. Regardless of its region's scale, each unit carries out identical operations at each iteration, and these operations happen in parallel at each scale. By sharing information through sparse connections between parents and children, the units collaborate to produce a consistent scene representation over the image plane. From an implementation perspective, this structure allows estimation to be trivially parallelized across multiple cores and single instruction multiple data (SIMD) channels. Experiments on the binocular stereo problem show that the consensus framework achieves greater accuracy on the KITTI benchmark~\cite{kitti} than comparable state-of-the-art variational and MRF approaches.

\section{Related Work}

There are many techniques for low-level vision problems like binocular stereo, optical flow, and shape-from-shading.  While they vary greatly in the way they derive information point-wise from image cues, their mechanisms for spatial aggregation tend to follow one of three different paradigms. The simplest paradigm is purely local---a single support region is explicitly defined around each pixel~\cite{kanade1994stereo,cvfilt1,cvfilt2,prs,yoon2006adaptive}. These regions are typically determined using intensity and texture information, either independently for each pixel or jointly for all pixels via segmentation, and they succeed when color and texture boundaries are well aligned with boundaries in the latent scene map.

Variational methods form another category. Estimation involves minimizing a per-pixel data cost along with a spatial regularization term that penalizes large derivatives in the scene map~\cite{bredies2010total,bruhn2006multigrid, horn1986variational,kuschkfast,ranftl2013minimizing,slesareva2005optic}. The derivative filters are designed to measure deviations from some implied local model, and the penalty is chosen to promote piecewise adherence while still being convex. Some variational methods employ multi-scale reasoning, through sequential coarse-to-fine optimization~\cite{bruhn2006multigrid} or simultaneous penalization of derivatives at multiple scales~\cite{BarronMalikTR2013}.

The third dominant paradigm are MRF-based  methods~\cite{lempitsky2008fusionflow, sun2003stereo, woodford2009global, PCBP,yamaguchi2013robust,SPS-St}. These explicitly encode piece-wise adherence to the local model (as opposed to the convex penalties in variational methods, which do so implicitly), by making hard decisions about the local model being valid across an edge or clique. Since they often consider continuous label spaces and non-submodular smoothness terms, these methods tend to rely on expensive approximate algorithms for optimization. Computation can be reduced by defining graphs on super-pixels instead of pixels~\cite{PCBP,yamaguchi2013robust,SPS-St}, and this does not substantially reduce accuracy as long as the super-pixel boundaries happen to be well aligned with scene boundaries.

The consensus framework is different from traditional, single-scale MRF techniques because it is defined on overlapping regions at multiple scales. It is also different from multi-scale MRF formulations that have been used for segmentation~\cite{pylon}, where parent nodes encode semantic context for co-occurring labels of their children. In consensus, all regions at all scales are self-similar. They all make direct predictions about pixel-level scene values, and they all use the same local model.

Consensus is inspired by our previous work on shape-from-shading~\cite{qsfs}. The objective in that paper can be seen as a special case of the one proposed here. Here, we introduce an encoding of local models that broadens the approach to a variety of low-level vision tasks. We also show that this encoding, when combined with a hierarchical organization that can be applied to a broad class of region sets, dramatically reduces computational expense by sharing information between parent and child regions in the hierarchy.

We use an alternating algorithm to minimize our objective. This is similar to ``divide and concur'' optimization algorithms like the \emph{alternating direction method of multipliers} (ADMM)~\cite{admm} that modify an objective to create multiple copies of a variable---one for each term in the original objective that includes that variable---and then enforce consistency between these copies. Our consensus objective resembles these modified, split objectives. A crucial part of our approach is the hierarchical organization of regions across scales, which makes the aggregation steps in the alternating minimization tractable. It is worth noting the approach of \cite{densecrf} here, which also uses an efficient data-structure for message aggregation during mean-field inference in a densely connected graph.

\section{Framework Elements}
\label{sec:body1}

We begin with a formal description of the three main components of the proposed framework. First, there is the global scene map. This is a function $Z(n) \in \mathbb{R}^d$ on the two-dimensional image plane, with $n=(x,y)$ indexing discrete spatial locations. $Z(n)$ may be scalar-valued ($d=1$) for properties such as stereo disparity, or  vector-valued for properties such as motion and 3D surface orientation.

Second, there is a dense set $P$ of overlapping regions $p \in P$ within the image plane, each one a collection of locations $n$. Set $P$ has regions at $K$ different scales, and symbol $P_k$ represents the subset of regions at scale $k\in\{1\ldots K\}$.  By convention, larger values of $k$ correspond to larger regions. Moreover, the regions can be organized hierarchically: for every region $p \in P_k$ at scale $k > 1$, it is possible to select a set $H_p$ of non-overlapping ``child regions'' from scales smaller than $k$, such that $p$ can be written exactly as their disjoint union. Figure.~\ref{fig:fwork}~(c,d) shows an example of such a region set for a one-dimensional image plane, where each $P_k$ is the set of overlapping regions of length $2^k$, and each $p\in P_k, k > 1$ is the union of two children from $P_{k-1}$. 

The final component is the local model. It is expected to apply piecewise across most of the scene, and it restricts accordingly the allowable choices for  scene values within any region $p$. It is encoded in a  mathematical form that encompasses all sorts of local models proposed in computer vision~\cite{planar,qsfs}, while also enabling the system to exploit the computational redundancy inherent to a hierarchy of regions. The local model says that scene values within any region $p$ must satisfy:
\begin{equation}
  \label{eq:modeldef}
  Z(n)=U(n)\theta_p,\qquad\forall n\in p, 
\end{equation}
where $U(n)\in \mathbb{R}^{d\times M}$ is some pre-defined matrix-valued function on the image plane, and $\theta_p \in \mathbb{R}^M$ is a variable associated with region $p$. Algebraically, this restricts local scene values to an $M$-dimensional linear sub-space, regardless of region size; and as a consequence of using a common $U(n)$, local scene estimates from two overlapping regions $p$ and $p'$ agree whenever $\theta_p=\theta_{p'}$. Here are some examples of functions $U(n)$ and their corresponding physical interpretations:
\begin{equation*}\label{eq:uexamples}
\begin{array}{ll}
U(n) =&[x~~y~~1], d = 1, M = 3,\\&\mbox{(disparity of locally-planar surfaces),}\vspace{1em}\\
U(n) = &\left[\begin{array}{c}\partial/\partial x\\\partial/\partial y\end{array}\right][x^2~~y^2~~xy~~x~~y], d = 2, M = 5,\\&\mbox{(normals of locally-quadratic surfaces),}\vspace{1em}\\
U(n)=&\left[\begin{array}{cc}x~~y~~1&0\\0&x~~y~~1\end{array}\right], d = 2, M = 6,\\
&\mbox{(flow vectors for locally-affine motion).}
\end{array}
\end{equation*}

With the three components in hand, estimation requires determining: a) which regions $p \in P$ are inliers with respect to the local model; and b) for all inlying regions, values of the per-region variables $\theta_p$ that are supported by the image data and consistent with each other. Inliers are indicated by a binary variable $I_p\in\{0,1\}$ associated with each patch.
Once determined, the values of $\{I_p,\theta_p\}$ together provide a rich and over-complete representation of the physical scene. At each point $n$, local grouping information is available through the subset $J_n$ of (potentially thousands of) inlying regions covering that point:
\begin{equation}
  \label{eq:jndef}
J_n = \{p\!\!: p \ni n, I_p = 1\}.  
\end{equation}
An estimate $\bar{Z}$ of the global scene map is induced as the point-wise average, or \emph{consensus}, of the local estimates from inlying regions:
\begin{equation}
  \label{eq:consensus}
\bar{Z}(n) = \frac{1}{|J_n|}\sum_{p \in J_n} U(n)\theta_p = \frac{1}{\sum_{p\ni n}I_p} \sum_{p\ni n} U(n)\theta_p~I_p.
\end{equation}
The count $|J(n)|$ represents the \emph{degree of consensus} at each point, and provides a point-wise measure of confidence in the estimate $\bar{Z}$.

Estimation is then cast as a minimization of the following cost over variables $\{I_p,\theta_p\}$:
\begin{align}
  L(\{I_p,\theta_p\}_{p\in P}) =&\sum_{p:I_p = 0} \tau_p + \sum_{p:I_p = 1} D_p(\theta_p)\notag\\&+ \lambda \sum_n~|J_n|~~\mbox{Var}\left[ \left\{U(n)\theta_p\right\}_{p \in J_n} \right].
  \label{eq:maincost}
\end{align}
The first term applies a cost $\tau_p$ for declaring region $p$ an outlier, in line with intuition that the local model is often valid. The second term scores local variables $\theta_p$ in each inlying region using \emph{data cost} $D_p(\cdot)$,  typically measuring the ability of restricted local scene estimates $U(n)\theta_p,\forall n\in p$ to explain the relevant image data. Both $\tau_p$ and $D_p(\cdot)$ can optionally be augmented to encode prior information about the scene or context from semantic visual processes. The final $\lambda$-weighted term promotes consistency between overlapping regions by penalizing, at every point, the variance of the scene predictions from inlying regions that cover it.
 
\section{Optimization Algorithm}
\label{sec:body2}

To minimize \eqref{eq:maincost},  we re-write the consistency term in terms of the global scene map $Z$, creating a related cost $L'$:
\begin{align}
    L'(\{I_p,\theta_p\}_{p\in P},Z) = \sum_{p:I_p = 0} \tau_p + \sum_{p:I_p = 1} \bigg( D_p(\theta_p)\notag\\+\lambda \sum_{n\in p} \|U(n)\theta_p-Z(n)\|^2\bigg),
    \label{eq:cost2}
\end{align}
where the two costs are equal when $Z$ is set to the consensus, \ie $L'(\{I_p,\theta_p\},\bar{Z}) = L(\{I_p,\theta_p\})$. In this new cost, the per-region summations are quadratic functions of variables $\theta_p$:
\begin{align}
 C_p(\theta_p,Z) = \sum_{n\in p} \left\|U(n)\theta_p - Z(n)\right\|^2 \notag\\
 =  \theta_p^TQ_p\theta_p -2\phi_p^T\theta_p + e_p,
  \label{eq:pcost}
\end{align}
with each $Q_p = \sum_{n\in p} U(n)^TU(n)$ a pre-computed $M\times M$ matrix  permanently associated with region $p$; and  each $\phi_p,e_p$ an $M$-vector and a scalar, respectively, derived from $Z$ as:
\begin{align}
 \phi_p = \sum_{n\in p} U(n)^TZ(n),\qquad e_p = \sum_{n\in p}\|Z(n)\|^2.
 \label{eq:phiedef}
\end{align}

Cost $L'$ is minimized iteratively, with each iteration having two steps. The first step is a minimization over region variables $\{I_p,\theta_p\}$ with  $Z$ fixed. Conveniently, this can be done independently---and in parallel---for each region since there are no cross-region terms in $L'$ when $Z$ is fixed. These independent minimizations are achieved by setting
\begin{equation}
  \label{eq:mintheta}
  \theta_p = \arg \min_\theta [D_p(\theta)+\lambda C_p(\theta,Z)],  
\end{equation}
and then, 
\begin{equation}
  \label{eq:mini}
  I_p = \left\{\begin{array}{cl}
       0, & \mbox{~if~} [D_p(\theta_p) + \lambda C_p(\theta_p,Z)] > \tau_p,\\
       1, & \mbox{otherwise .}
       \end{array}\right.
\end{equation}
In other words, the best model-based explanation is found for each region $p$, and then the region is declared outlier if the error-of-fit exceeds the outlier cost $\tau_p$. 

The second step at each iteration is a minimization over $Z$ with region variables fixed at their new values. This is achieved simply by setting $Z=\bar{Z}$ as per \eqref{eq:consensus}, and it is thus guaranteed that $L(\{I_p,\theta_p\})=L'(\{I_p,\theta_p\},Z)$ at the end of every iteration.
Consequently, beginning with any initial estimate of the scene map $Z$, each iteration decreases the value of $L'$, and therefore of $L$, which converges to a (local) minimum whenever $\{D_p(\cdot)\}$ have finite lower bounds. 

Convergence to a good local minimum is promoted by beginning the iterations with a smaller value for the consistency weight $\lambda$, and then increasing it to its final value across the initial iterations. Interestingly, this induces a temporal coarse-to-fine refinement of the scene map during the optimization. Early-on, smaller $\lambda$ values allow more inlying regions, causing the consensus to be smoothed across larger areas. As $\lambda$ increases, more regions that span scene discontinuities become outliers, and the consensus exhibits progressively finer detail.

\subsection{Hierarchical Computation}

The computational cost of this optimization depends on the complexities of the three parts of every iteration: 
\begin{enumerate}[topsep=0ex,itemsep=-1ex]
\item Computing  intermediate regional consistency terms \{$\phi_p,e_p$\} from $Z$ as per \eqref{eq:phiedef}.
\item Updating $(\theta_p, I_p)$ for every region $p$ as per \eqref{eq:mintheta},\eqref{eq:mini}.
\item Setting $Z = \bar{Z}$ as per \eqref{eq:consensus}.
\end{enumerate}

The complexity of the second part depends on the  forms of functions $D_p(\cdot)$, but it usually scales well because it involves parallel optimization over $M$-dimensional domains (regardless of region size). The first and third parts would scale poorly if implemented naively, but this can be averted by exploiting the hierarchical structure of $P$.

First, consider the computation of $\{\phi_p,e_p\}$ from $Z$. Using the fact that every region $p$ is partitioned by its child regions $c\in H_p$, we can write
\begin{align}
  \phi_p = \sum_{n \in p} U(n)^TZ(n)
  = \sum_{c \in H_p} \sum_{n\in c} U(n)^TZ(n) 
  = \sum_{c \in H_p} \phi_c,
  \label{eq:upsweep}
\end{align}
and similarly, $e_p = \sum_{c\in H_p} e_p$. This reduces the number of additions significantly---from the size of the region $p$ to just the number of its children. To ensure that values of $\{\phi_c, e_c\}_{c\in H_p}$ are available,  calculations of $\{\phi_p,e_p\}$ are scheduled in an upward sweep through the hierarchy, using explicit summation over $n$ for regions at scale $k=1$, and the cheaper right-most expression of \eqref{eq:upsweep} for progressively larger scales.

The hierarchical structure can also be leveraged to efficiently compute the consensus $\bar{Z}$ from the current values of the region variables $\{\theta_p,I_p\}$. Note that for every region $p\ni n$ at scale $k > 1$, there is one and only one child region in $H_p$ that also includes $n$. For the simple case with only two scales ($K=2$), we see that the summation of local estimates from inlying regions can be simplified to
\begin{align}
  \label{eq:downsimp}
  \sum_{\{p \ni n\}\cap P_1} U(n)\theta_pI_p + \sum_{\{p \ni n\}\cap P_{2}} U(n)\theta_pI_p\notag\\
  = \sum_{\{p \ni n\}\cap P_1} U(n)\left(\theta_pI_p + \sum_{r \in H_p^{-1}} \theta_rI_r\right),
\end{align}
where $H^{-1}_p = \{r: H_r \ni p\}$ denotes the set of \emph{parents} for any region $p$\footnote{Note that for a region $p \in P_K$ at the largest scale, $H^{-1}_p=\emptyset$.}. In the more general case with $K$ scales, we recursively define augmented variables $\{\theta^+_p,I^+_p\}$ for every region $p$ as
\begin{equation}
  \label{eq:downrec}
  \theta^+_p = \theta_p~I_p + \sum_{r\in H_p^{-1}} \theta^+_r,~~~I^+_p = I_p + \sum_{r\in H_p^{-1}}I^+_r,
\end{equation}
which can be computed by a \emph{downward} sweep through the pyramid. Then, it is easy to see that the numerator and denominator of the expression for $\bar{Z}(n)$ in \eqref{eq:consensus} are given by
\begin{align}
  \left(\sum_{p\ni n}U(n)\theta_p~I_p\right)&= U(n) \sum_{\{p\ni n\}\cap P_1} \theta^+_p,\notag\\
  \left(\sum_{p\ni n}I_p\right) &= \sum_{\{p\ni n\}\cap P_1}I^+_p.
\end{align}
Thus, instead of computing summations over all overlapping regions at all scales for each location $n$, the consensus can be computed using summations over the augmented variables $\{\theta_p^+,I_p^+\}$ of regions at just the smallest scale. 

The gains from using these recursive computations is substantial, and can be interpreted as reducing the \emph{effective} connectivity of the framework to just the sparse set of hierarchical links. For the network in Fig.~\ref{fig:fwork} (c,d), it represents a reduction, in the number of required summations for \eqref{eq:consensus} and \eqref{eq:phiedef}, from $\mathcal{O}(2^{K}N)$ to $\mathcal{O}(KN)$. Moreover, while the recursion requires different scales to be processed sequentially, note that the computations in \eqref{eq:upsweep} and \eqref{eq:downrec} can still be carried out for all regions $p \in P_k$ at each scale $k$ in parallel. 

Therefore, as visualized in Fig.~\ref{fig:fwork}~(d), computation happens in a distributed architecture, requires the identical operations of \eqref{eq:mintheta},\eqref{eq:mini},\eqref{eq:upsweep}, and \eqref{eq:downrec} at each region, with operations at each scale happening in parallel and information being passed through hierarchical links between scales---all of which arises naturally as an efficient way to optimize a well-defined mathematical objective. A complete listing of the algorithm is included in the supplementary material.

\section{Application to Stereo }

In this section, we describe the application of the consensus framework to the task of stereo estimation, and its evaluation on the KITTI benchmark~\cite{kitti}. We reason with a planar local model (\ie, $U(n)=[x~~y~~1]$), and define our region set $P$ to be a two-dimensional equivalent of the pyramid in Fig.~\ref{fig:fwork}(b,c), with five scales consisting of all overlapping patches of sizes $4\times 4$, $8\times 8$, $16\times 16$, $32\times 32$, and $64\times 64$, where patches at all but the finest scale can be partitioned into four children from the next lower scale.

We follow the approach of \cite{PCBP,yamaguchi2013robust,SPS-St} in defining our data cost $D_p(\cdot)$. In particular, we use the implementation from Yamaguchi et al.~\cite{SPS-St}, that implements semi-global matching (SGM)~\cite{SGM} with a cost based on absolute differences of gradients and the Census transform~\cite{census}, to compute an initial set of approximate disparity estimates $Z_{\mbox{\tiny SGM}}(n)$ at a semi-dense set of locations $n \in \Omega_{{\tiny SGM}}$. The data costs for every region $p$ are then defined as:
\begin{equation}
  \label{eq:dpdef}
  D_{p}(\theta) = \sum_{n \in p} w_{\mbox{\tiny SGM}}(n) \bigg( U(n)\theta - Z_{\mbox{\tiny SGM}}(n) \bigg)^2,
\end{equation}
where $w_{\mbox{\tiny SGM}}(n) = 0$ if $n \notin \Omega_{\mbox{\tiny SGM}}$, $1/4$ if there is a discontinuity in $Z_{\mbox{\tiny SGM}}$ around $n$, and $1$ otherwise. Note that since each $D_p(\cdot)$ above is also a quadratic function of $\theta$, the minimizations in \eqref{eq:mintheta} simply involve solving a $3\times 3$ linear system for each region (and moreover, we can re-use the results of LDL decompositions across iterations when $\lambda$ is constant).

The scalar outlier costs $\{\tau_p\}$ for various patches are defined to be proportional to their size $|p|$ as:
\begin{equation}
  \label{eq:taudef}
  \tau_p = \tau_0 \times |p| \times \max\left(0.5,\exp\left(-0.25V_p^2\right)\right).
\end{equation}
Here, $V_p$ measures if $p$'s children are a better fit to a different parent based on intensity variance, and is the count of the number of patches that share a child with $p$ and whose intensity variance is lower than that of $p$. Through cross-validation, we set $\tau_0=1.44$ and the consistency weight $\lambda = 0.4$, and as described in Sec.~\ref{sec:body2}, we begin the iterations with a lower value of $\lambda' = 2^{-18}\lambda$, and increase it by a factor of eight every six iterations till it reaches its final value. 

\begin{figure*}
  \centering
  \includegraphics[width=\textwidth]{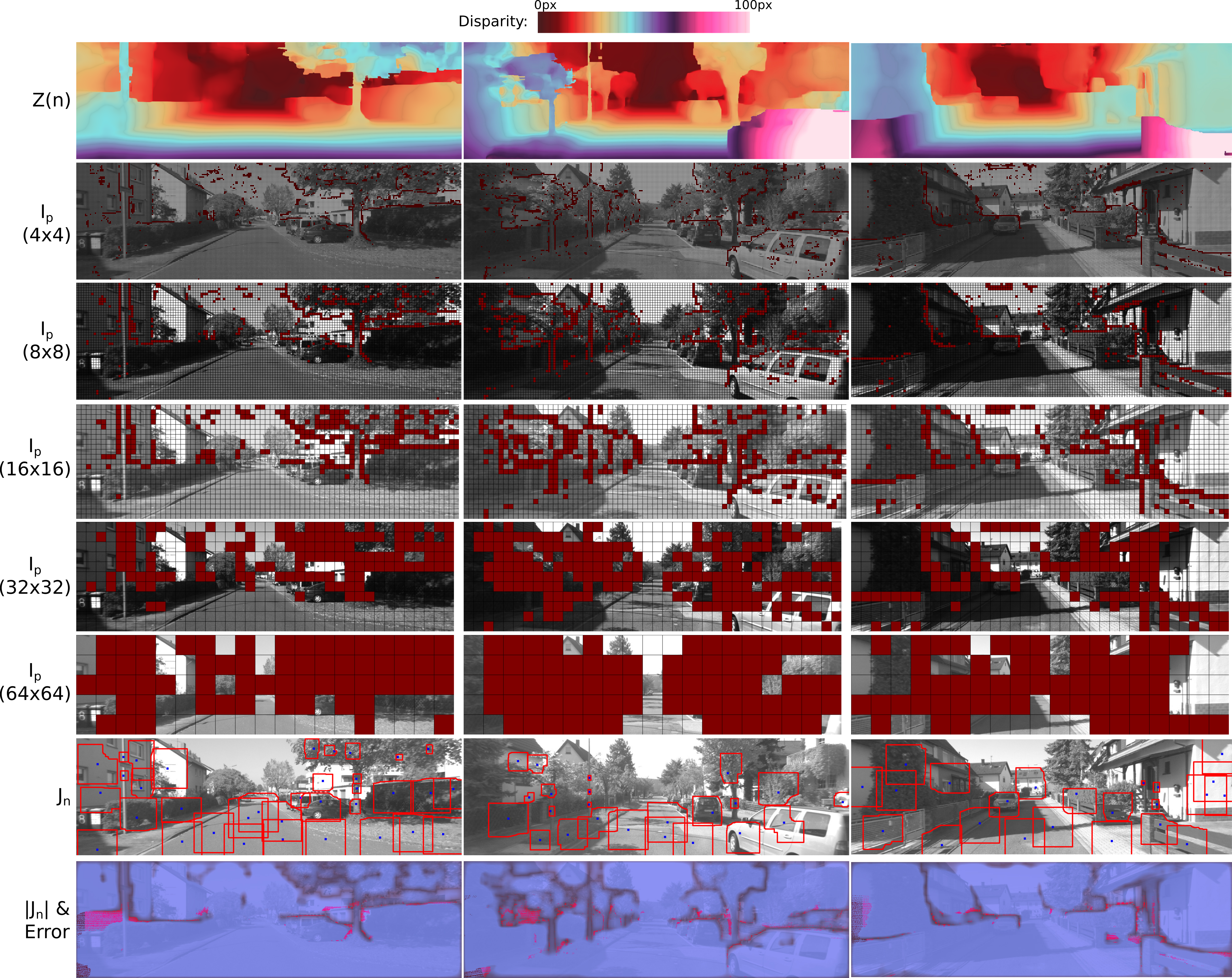}
  \caption{Framework output for three image pairs from the KITTI training set. {\bf (Row 1)} Scene map formed through consensus of predictions from all inlying regions. {\bf (Row 2-6)} Inlier statuses of regions at different scales, superimposed on the left images of each stereo pair. For clarity, we only show the statuses of a non-overlapping subset of regions at each scale. {\bf (Row 7)} Boundaries (in red) of the support region for various points $n$ (in blue), formed as the union of their inlying consensus set $J_n$. {\bf (Row 8)} Degree of consensus $|J_n|$ (blue saturation) and sites of erroneous estimates (red), defined as estimates with error greater than 3 pixels. (Erroneous estimates whose ground truth disparities place them outside the field of view of the right camera are shown as dark red.)}
  \label{fig:results}
\end{figure*}

Finally, we add a way to incorporate reasoning about occlusions. While one could achieve this encoding using more sophisticated definitions of $D_p(\cdot)$ or $\tau_p$, we find that a much simpler approach suffices. We use the fact that the pixels missing in $\Omega_{\mbox{\tiny SGM}}$ correspond to those that have failed a left-right consistency check, and are likely occluded. Using the data costs defined as above, the consensus framework usually yields a scene map where disparity values in occluded regions, in the absence of any input data, are interpolated between the occluded and occluding planes either smoothly, or with a discontinuity at an arbitrary location within the region. In order to incorporate the intuition that occluded pixels are likely to be part of the occluded background, we run the alternating minimization for fifty iterations; then set the values of the consensus $Z(n)$, at the potentially occluded points in $n\notin \Omega_{\mbox{\tiny SGM}}$, to the lower of their current value and that of the background pixel on the same epipolar line (\ie, the nearest pixel on the same horizontal line in $\Omega_{\mbox{\tiny SGM}}$); and then run the minimization for another thirty iterations.

A reference implementation (available on the project page), designed to make use of thread-based parallelism and SIMD instruction sets, takes an average of only six seconds (1.5 seconds for the initial SGM step) on a $1240\times 370$ image, when running on a CPU with six cores. Moreover, since the computations in the framework have a degree of parallelism roughly equal to the resolution of the input image, we expect execution time will continue to decrease with access to larger numbers of cores.

\subsection{Evaluation on KITTI}

We evaluate the proposed algorithm on the KITTI benchmark~\cite{kitti} which contains a total of 389 grayscale image pairs of rural road scenes, captured using an autonomous driving platform equipped with a pair of high-resolution cameras.  A Velodyne laser scanner provides ground truth at a subset of pixels in each scene. This ground truth is made available for a subset of 194 image pairs---the \emph{training set}---and withheld for the remaining image pairs that form the \emph{testing set}. A website associated with the database tracks the performance of stereo algorithms on the testing set. Note that while the benchmark also contains temporally-adjacent stereo frames that allow simultaneous reasoning about optical flow and stereo, we ignore those extra frames and consider the pure stereo problem here.

Figure~\ref{fig:results} visualizes various aspects of the internal representation of our framework on convergence, for three scenes in the KITTI training set. The top row shows the consensus global disparity map, and Rows 2--6 visualizes a regularly-spaced subset of in the inlier statuses $I_p$. Row 7 provides another view of variables $I_p$, by explicitly showing some of the  ``support regions'' formed as the union of all patches in $J_n$, for various pixels $n$. These regions by-and-large group together points whose disparity values would be well-explained by a slanted plane model. As expected, there is significant variation in the size and shapes of the support regions across each scene, matching the scale of the underlying scene structures. This highlights the distinction from superpixel-based MRF approaches~\cite{PCBP,yamaguchi2013robust,SPS-St}, which require choosing a single scale for the entire scene. Also note that for many pairs of points that do not directly lie in each others' support regions, the regions themselves have significant overlap. Through such overlap, the consensus estimate at a point has benefited from aggregation across regions that are larger than the union of the set of patches that include it.

\begin{figure}[!t]
\centering
\includegraphics[height=14em]{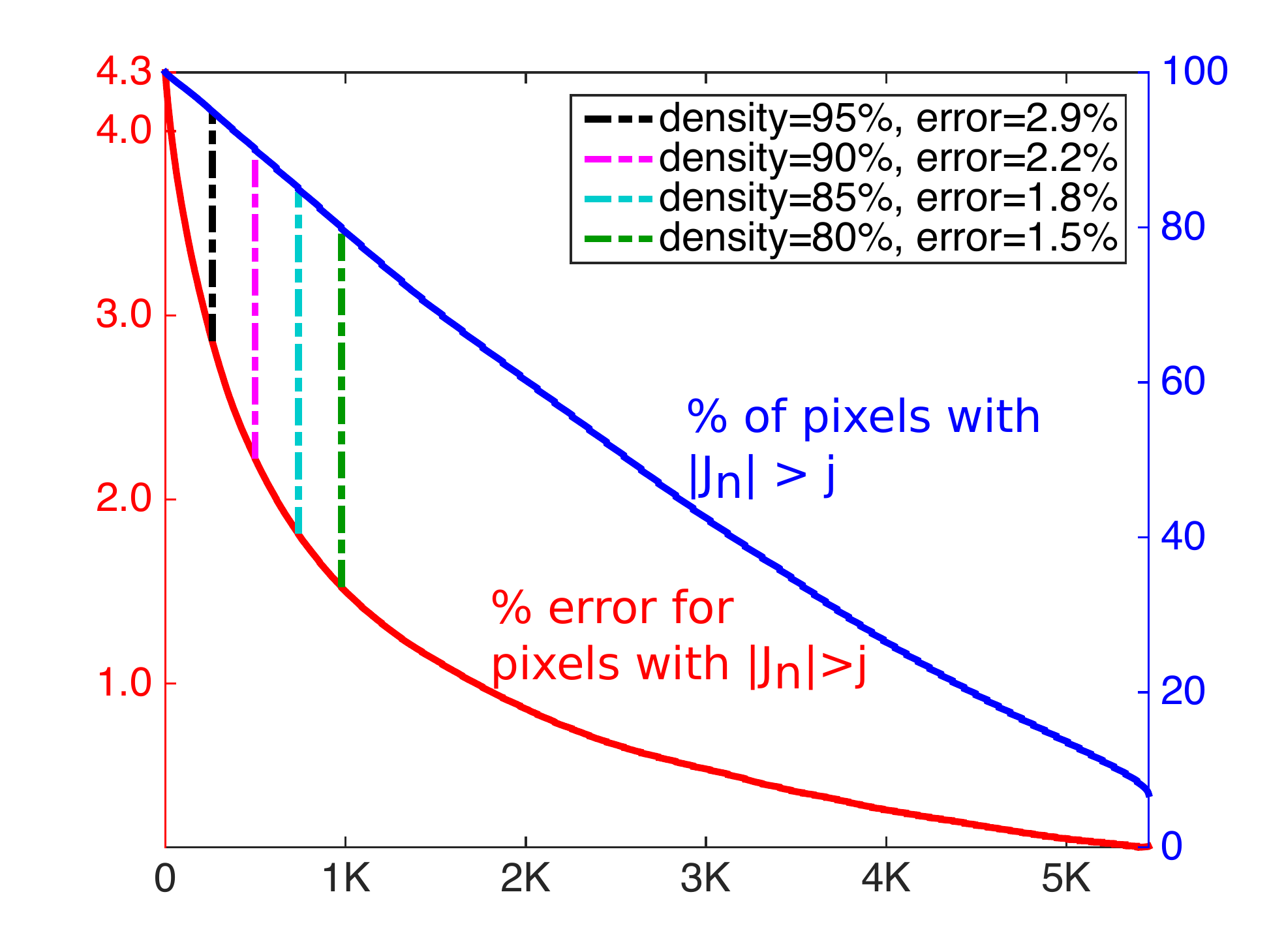}
\caption{Error vs. degree of consensus $|J_n|$. Blue curve shows percentage of points with $|J_n|$ above different thresholds, and red curve their corresponding error rate, in terms of percentage with error $> 3$ px. These are computed over all pixels with ground truth data available, across all images in the KITTI training set.}
\label{fig:jnerr}
\end{figure}

The final row in Fig.~\ref{fig:results} visualizes the degree of consensus $|J_n|$ at all points (blue saturation), simultaneously with locations of erroneous estimates (red). We see that many of the errors occur around object boundaries and near small scene structures, which are also points where $|J_n|$ is low. We quantify this observation in Fig.~\ref{fig:jnerr}, and find that average estimation error drops rapidly as we discard points with the lowest values of $|J_n|$. This an other benefit of the rich internal representation: In addition to providing a global scene estimate, it also provides a natural measure of point-wise confidence in this estimate.

\begin{figure*}[!t]
  \centering
  \begin{tabular}{ccccc}
    \includegraphics[width=0.19\textwidth]{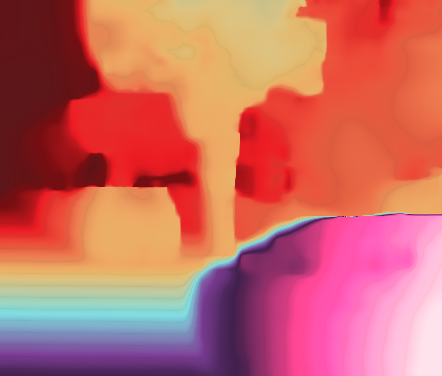}\hspace{-1em}~&
    \includegraphics[width=0.19\textwidth]{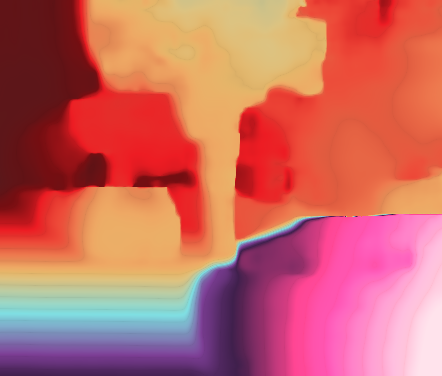}\hspace{-1em}~&
    \includegraphics[width=0.19\textwidth]{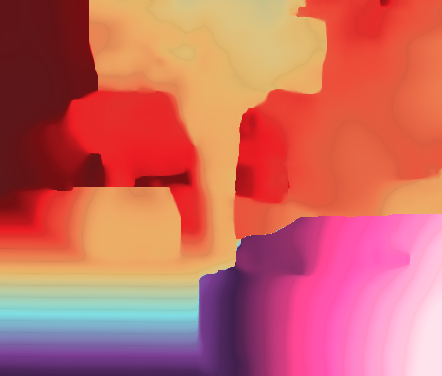}\hspace{-1em}~&
    \includegraphics[width=0.19\textwidth]{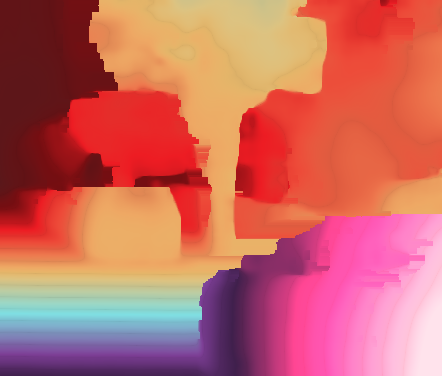}\hspace{-1em}~&
    \includegraphics[width=0.19\textwidth]{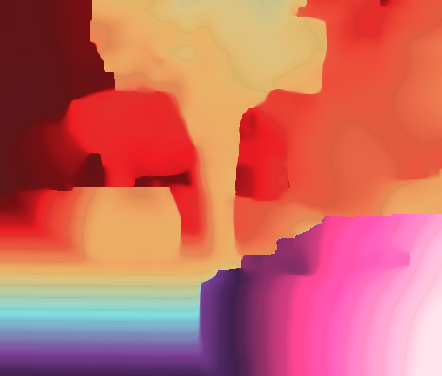}\vspace{-0.5em}\\
    \footnotesize Iteration \#15 & 
    \footnotesize Iteration \#30 & 
    \footnotesize Iteration \#50 & 
    \footnotesize ~~~~Occlusion-based Correction & 
    \footnotesize Iteration \#80 (Final)
  \end{tabular}\vspace{0.5em}
  \caption{Evolution of the scene map across iterations. The first two frames correspond to iterations when the consistency weight $\lambda$ is still being increased towards its final value, causing the map to undergo a coarse-to-fine refinement. After iteration \#50, we apply the occlusion-based correction step, and then run the minimization for another 30 iterations.}
  \label{fig:temporal}
\end{figure*}

Next, we visualize the progression of the alternating minimization algorithm. Figure~\ref{fig:temporal} shows the evolution of the consensus $\bar{Z}(n)$ across iterations. In the early iterations as the consistency weight $\lambda$ is increased toward its final value, the map undergoes a coarse-to-fine refinement, with image boundaries becoming sharper. After the fiftieth iteration, the occlusion correction is applied to propagate disparity estimates from background planes into potentially-occluded regions. Since this is a simple correction applied independently along each epipolar line, it introduces inconsistencies within occluded regions. The scene map therefore benefits from further refinement through another thirty iterations of minimizing the consensus cost, yielding a final estimate that is more accurate. The supplementary material includes additional results showing the evolution of the objective \eqref{eq:maincost} for different initial values and update schedules for $\lambda$.

\definecolor{First}{rgb}{0.45,0.55,1.0}
\definecolor{Second}{rgb}{0.7,0.69,1.0}
\definecolor{Third}{rgb}{0.7,0.75,0.9}

\newcommand{\sdc}{\textcolor{red}{*}}

\begin{table*}[!t]
  \centering
  {\small
  \begin{tabular}{|r||c|c||c|c||c|c||c|c||c|c||c|}
    \hline
    \multirow{2}{*}{Method} &
    \multicolumn{2}{c||}{Avg.~Error} &
    \multicolumn{2}{c||}{$>2$px} &
    \multicolumn{2}{c||}{$>3$px} &
    \multicolumn{2}{c||}{$>4$px} &
    \multicolumn{2}{c||}{$>5$px} & 
    \multirow{2}{*}{Exec.~Time}\\\cline{2-11}
    & All & NOC & All & NOC & All & NOC& All & NOC& All & NOC &\\\hline\hline

    ATGV~\cite{ranftl2013minimizing}
    \hspace{-0.8em}~& 1.6px & 1.0px & 9.05\% & 7.08\% & 6.88\% & 5.02\% & 5.76\% & 3.99\% & 5.01\% & 3.33\% & 6min: 8 cores\\\hline

    wSGM~\cite{wsgm}
    \hspace{-0.8em}~& 1.6px & 1.3px & 8.72\% & 7.27\% & 6.18\% & 4.97\% & 4.89\% & 3.88\% & 4.11\% & 3.25\% & 6s: 1 core\\\hline

    AARBM~\cite{aarbm}
    \hspace{-0.8em}~& 1.2px & 1.0px & 8.70\% & 7.36\% & 5.94\% & 4.86\% & 4.56\% & 3.67\% & 3.69\% & 2.96\% & 0.25s: 1 core\\\hline

    \sdc{}PCBP~\cite{PCBP}
    \hspace{-0.8em}~& 1.1px & 0.9px & 7.62\% & 5.08\% & 5.37\% & 4.04\% & 4.29\% & 3.14\% & 3.64\% & 2.64\% & 5min: 4 cores\\\hline

    \hspace{-0.6em}\sdc{}StereoSLIC~\cite{yamaguchi2013robust}
    \hspace{-0.8em}~&\cellcolor{Second}1.0px & 0.9px & 7.20\% & 5.76\% & 5.11\% & 3.92\% & 4.04\% & 3.04\% & 3.33\% & 2.49\% & 2.3s: 1 core\\\hline

    \sdc{}DDS-SS~\cite{dds}
    \hspace{-0.8em}~&\cellcolor{Second}1.0px & 0.9px & 6.96\% & 5.91\% & 4.59\% & 3.83\% &\cellcolor{Third}3.49\% & 2.90\% &\cellcolor{Third}2.83\% & 2.36\% & 1min: 1 core\\\hline

    \sdc{}PCBP-SS~\cite{yamaguchi2013robust} 
    \hspace{-0.8em}~&\cellcolor{Second}1.0px &\cellcolor{First}0.8px & 6.75\% & 5.19\% & 4.72\% & 3.40\% & 3.75\% &\cellcolor{Third}2.62\% & 3.15\% &\cellcolor{Third}2.18\% & 5min: 1 core\\\hline

    \sdc{}SPS-St~\cite{SPS-St} 
    \hspace{-0.8em}~&\cellcolor{Second}1.0px & 0.9px &\cellcolor{Third}6.28\% &\cellcolor{Third}4.98\% &\cellcolor{Third}4.41\% &\cellcolor{Third}3.39\% & 3.52\% & 2.72\% & 3.00\% & 2.33\% & 2s: 1 core\\\hline

    MC-CNN~\cite{mccnn} 
    \hspace{-0.8em}~&\cellcolor{Second}1.0px &\cellcolor{First}0.8px &\cellcolor{First}5.39\% &\cellcolor{First}4.30\% &\cellcolor{First}3.84\% &\cellcolor{First}2.61\% &\cellcolor{First}3.01\% &\cellcolor{First}2.04\% &\cellcolor{First}2.52\% &\cellcolor{First}1.75\% & 100s: GPU\\\hline\hline
    \sdc{}{\bf Proposed:} All $n$
    \hspace{-0.8em}~& \cellcolor{First}0.9px &\cellcolor{First}0.8px &\cellcolor{Second}5.88\% &\cellcolor{Second}4.85\% &\cellcolor{Second}4.10\% &\cellcolor{Second}3.30\% &\cellcolor{Second}3.26\% &\cellcolor{Second}2.59\% &\cellcolor{Second}2.74\% &\cellcolor{Second}2.16\% & \multirow{2}{*}{6s: 6 cores}\\\cline{2-11}
    \parbox[c]{6.3em}{\footnotesize\flushright\vspace{-0.5em}Only $|J_n| \geq 200$\\ (96.4\% density)}
    \hspace{-0.8em}~& 0.8px & 0.6px
    & 4.59\% & 3.50\%
    & 2.98\% & 2.14\%
    & 2.26\% & 1.56\%
    & 1.85\% & 1.24\% &\\
    \hline
  \end{tabular}}\vspace{0.5em}
  {\small \colorbox{First}{~~~Lowest~~~} ~~~ \colorbox{Second}{Second Lowest} ~~~ \colorbox{Third}{Third Lowest}} ~~~ \sdc{}: Same Matching Cost\vspace{0.5em}
  \caption{Comparison with the state-of-the-art on the KITTI testing set. Performance is measured in terms of average error, as well as percentage of estimates with error greater than different thresholds. For each metric, the ``All'' column reports values computed over all ground truth pixels, and ``NOC'' over only those those that are within the field-of-view of the right camera. The last row reports the accuracy of our method's estimates that have confidence measure $|J_n|$ above a threshold, and correspond to errors values computed over 96.4\% of the points with ground truth available.}
  \label{tab:res}
\end{table*}

Table~\ref{tab:res} compares the consensus framework with other state-of-the-art stereo algorithms\footnote{This table only includes methods that use just one stereo pair as input.} in terms of various error quantiles on the KITTI testing set. The most direct comparisons of our results are with those of \cite{dds,PCBP,yamaguchi2013robust,SPS-St}, since these methods all use the same approach  to derive their data costs (census transform and gradient-based matching with SGM). These only differ---from us, and from each other---in their approach to spatial aggregation. The consensus framework outperforms all of these methods on all error metrics, while also having a low execution time.

Table~\ref{tab:res} also reports the performance of the recently released MC-CNN~\cite{mccnn} algorithm, which computes point-wise matching costs using a multi-layer convolutional neural network. This produces lower error values than all other methods, including ours, in exchange for greater computation (the method takes 100 seconds on a GPU with 2880 CUDA cores). This is encouraging, because improved pixel-wise data costs like this one can be directly substituted into the consensus framework to enhance accuracy. 

Finally, we demonstrate the benefit of the pixel-wise confidence measure in our framework by reporting a second set of results in Table~\ref{tab:res}. This is simply produced by discarding a small number of pixels with the lowest degree of consensus $|J_n|$ (\ie those with degree less than $200$ out of the maximum possible value of $\sim 5500$). This second set of error quantiles---computed now on the high-confidence set with only $\sim 3.6\%$ fewer pixels---are the smallest of all methods. In this mode, the proposed algorithm efficiently produces very reliable disparity estimates, at all but a small fraction of locations. This also suggests a strategy for leveraging sophisticated matching strategies such as \cite{mccnn} when execution time is a bottleneck (such as for  automated driving applications)---one where the more expensive matching costs are computed only for the small number of low-confidence pixels.

\section{Conclusion}
\label{sec:conclusion}

In this paper, we introduced a framework for low-level visual estimation with local scene models that reasons with a  large overlapping, multi-scale set of regions, to determine which of them are outliers, and which of them can generate model-based scene value estimates while being consistent with each other. Despite the larger variable space, and the greater complexity of the consensus objective, we showed that optimization can be carried out efficiently by recognizing that the regions can be organized hierarchically. An evaluation on stereo estimation found that the framework outperforms existing approaches to spatial reasoning.

An important direction of future research lies in applying the framework to problems involving estimating different physical properties of the same scene (such as material and shape), with different piecewise local models for each, when the aggregation regions of one property suggest, but do not determine, those of the other.

On a different note, many properties of the consensus framework---multi-scale collaboration, implementation as a distributed architecture of computational units carrying out the same operations, coarse-to-fine evolution of the scene map, \etc---mimic behavior observed in biological systems~\cite{funcon}. It would be interesting to explore these links systematically---to investigate whether the framework, or some variation of it, can serve as a faithful model for biological processing; as well as whether insights from biology can be used to further improve the framework.

{\small
\subsection*{Acknowledgments}
YX and TZ acknowledge support from the NSF under award no.\,IIS-1212928.

}

\onecolumn

{\flushleft \bf \large Supplementary Material}~\vspace{0.6em}\\
{~}~\vspace{0.4em}

\small
\section*{Appendix A: Inference Algorithm Listing}
We present here a summary of the inference algorithm for reference. It takes as input the following elements:
{\parskip0pt \begin{enumerate}
\parskip0pt
\item Sets of regions $P_k, k \in \{1,\ldots K\}$ at $K$ different scales. 
\item For each region $p \in P_k, k > 1$, a set $H_p$ consisting of non-overlapping child regions that partition $p$, and are from scales smaller than $k$ (\ie $H_p \subset \bigcup_{k'=1}^{k-1}P_{k'}$).
\item The data cost functions $D_p(\cdot)$ and scalar outlier costs $\tau_p$ for every region $p$.
\item The value of the consistency weight $\lambda$. Additionally, its value $\lambda_0$ to be used at the beginning of the iterations, the factor $\lambda_f > 1$ by which it is to be increased at every $T_\lambda$ iterations.
\item An initial estimate $Z_0(n)$ of the scene value map.
\end{enumerate}}
\noindent Given these elements, the algorithm to minimize the cost function $L$ is reproduced below:

{\flushleft \parskip0pt
\hrule\vspace{0.25em}

\emph{\# Initialization}\\
Set $\lambda^* = \lambda_0, Z(n) = Z_0(n)$ for all $n$.\\
In parallel, {\bf for} all $p \in P_1$:\\
~~~~Set $Q_p = \sum_{n\in p}U(n)^TU(n)$.\\
{\bf end;}\\
{\bf for} $k = 2\ldots K$\\
~~~~In parallel, {\bf for} all $p \in P_k$:\\
~~~~~~~~Set $Q_p = \sum_{c \in H_p}Q_c$.\\
~~~~{\bf end;}\\
{\bf end;}\\~\\
\emph{\# Main Iterations}\\
{\bf for} iters = $1\ldots$MAXITERS\\
~~~~\emph{\# Upsweep}\\
~~~~In parallel, {\bf for} all $p \in P_1$:\\
~~~~~~~~Set $\phi_p = \sum_{n\in p} U(n)^TZ(n), e_p = \sum_{n\in p}\|Z(n)\|^2.$\\
~~~~{\bf end;}\\
~~~~{\bf for} $k = 2\ldots K$\\
~~~~~~~~In parallel, {\bf for} all $p \in P_k$:\\
~~~~~~~~~~~~Set $\phi_p,~~e_p$ as per (10).\\
~~~~~~~~{\bf end;}\\
~~~~{\bf end;}\\~\\
~~~~\emph{\# Minimize}\\
~~~~In parallel, {\bf for} all $p \in P$:\\
~~~~~~~~Set $\theta_p,I_p$ as per (8) and (9).\\
~~~~{\bf end;}\\~\\
~~~~\emph{\# Downsweep}\\
~~~~{\bf for} $k = K,K-1,\ldots 1$\\
~~~~~~~~In parallel, {\bf for} all $p \in P_k$:\\
~~~~~~~~~~~~Set $\theta^+_p,I^+_p$ as per (11).\\
~~~~~~~~{\bf end;}\\
~~~~{\bf end;}\\
~~~~In parallel, {\bf for} all $n$:\\
~~~~~~~~Set $Z(n)$ as per (12).\\
~~~~{\bf end;}\\~\\
~~~~\emph{\# Update $\lambda^*$}\\
~~~~Set $\lambda^*=$MIN$(\lambda^* \times \lambda_f, \lambda)$ {\bf if} mod(iters,$T_\lambda$) = 0.\\
{\bf end;}\\
\hrule\vspace{0.25em}
}
.
\section*{Appendix B: Evolution of Consensus Objective during Optimization}
\begin{figure}[!h]
\centering
\includegraphics[width=0.88\textwidth]{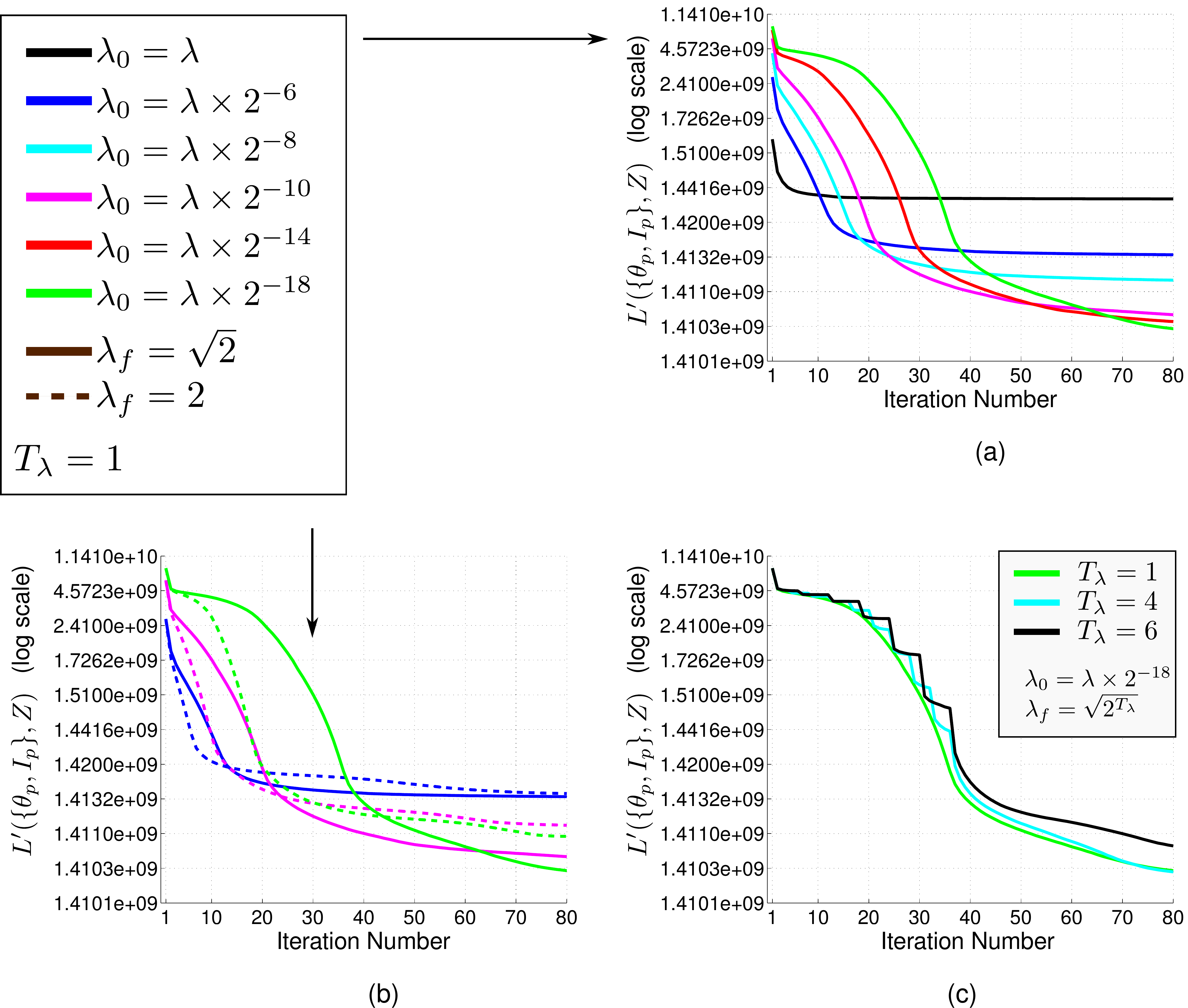}
\caption{This figure shows the evolution of the consensus cost during optimization for a typical image with using different initial values $\lambda_0$ and update schedules $\lambda_f, T_\lambda$ (see Appendix A) for $\lambda'$. The consensus cost shown is computed with the true value of the consistency weight $\lambda$ (even for the iterations when minimization is done with lower values $\lambda'$), and the occlusion-based correction step is omitted.}
\label{fig:cevol}
\end{figure}

As described in Sec.~\ref{sec:body2}, to avoid poor local minima and promote convergence to a good solution with a low cost, we use a lower value $\lambda'$ of the consistency weight in the early iterations of the alternating minimization method, and increase it slowly to the desired weight $\lambda$. Figure \ref{fig:cevol} illustrates the effect of different schedules for $\lambda'$ on convergence for a typical example. In Fig.~\ref{fig:cevol}~(a), we show the evolution of the objective starting with different values of $\lambda'=\lambda_0$, and increasing it by a constant factor of $\lambda_f=\sqrt{2}$ at every iteration, and keeping it fixed after it reaches $\lambda'=\lambda$. We see that the direct alternating minimization case ($\lambda_0=\lambda$) decreases the consensus cost sharply in the first few iterations, but then stagnates at a local minima with a relatively high cost. As we lower the starting value of $\lambda_0$, the cost has higher values and decreases more gradually in the initial iterations, but continues to decrease over a larger number of iterations and eventually converges to a better solution with a lower cost. Figure~\ref{fig:cevol}~(b) explores the effect of a higher rate $\lambda_f$ of increasing $\lambda'$. We see that like with a lower starting value for $\lambda_0$, a slower rate $\lambda_f$ leads to convergence to a better solution, albeit more gradually.

In addition to requiring more iterations to converge, another computational penalty of changing $\lambda'$ across iterations is that it requires re-doing any pre-computations that depend on the consistency weight. For our stereo algorithm, minimizing the sum of the data and consistency costs involves solving a $3\times 3$ linear system for each region, and changing the value of $\lambda'$ requires re-doing the LDL decompositions of the system matrices. Since this is expensive, it is desirable to avoid changing the value of $\lambda'$ at every iteration. In Fig.~\ref{fig:cevol}~(c), we consider different cases with the same value of $\lambda_0$ while jointly setting the increase factor $\lambda_f$, and the interval $T_\lambda$ at which it is applied, so that the total number of iterations taken for $\lambda'$ to reach its final value $\lambda$ remains the same (\ie, by applying a higher rate at larger intervals). We see that choosing higher intervals leads to a ``stair-casing'' effect in the evolution of the objective, but the solution it converges to is only worse by a relatively small margin. We find this to be an acceptable trade-off between convergence to a low-cost solution and limiting computational expense, and use the parameters $\lambda_0=\lambda / 2^{18}, \lambda_f = \sqrt{2^{T_\lambda}}, T_\lambda = 6$ in our stereo implementation.

\end{document}